\pdfoutput=1

\documentclass[11pt]{article}

\usepackage[final]{acl}

\usepackage{times}
\usepackage{latexsym}
\usepackage{enumitem}
\usepackage[T1]{fontenc}

\usepackage[utf8]{inputenc}
\usepackage{xcolor}
\usepackage{pifont}
\usepackage{microtype}

\usepackage{inconsolata}

\usepackage{graphicx}

\usepackage{amsmath} 
\usepackage{booktabs}
\usepackage{multirow} 
\usepackage{adjustbox} 
\usepackage{subcaption}  
\usepackage{makecell}    

\title{Context-Fidelity Boosting: Enhancing Faithful Generation through Watermark-Inspired Decoding
}

\author{%
  Weixu Zhang$^{1,2,3}$$^{\ast}$, Fanghua Ye$^{1}$$^{\ast}$, Qiang Gao$^{1,4}$, Jian Li$^{1}$$^{\dagger}$, Haolun Wu$^{2,3}$,\\  \textbf{Yuxing Tian$^{5}$, Sijing Duan$^{6}$, Nan Du$^{1}$, Xiaolong Li$^{1}$, Xue Liu$^{2,3,7}$}\\
  $^1$Hunyuan AI Digital Human, Tencent 
  $^2$McGill University 
  $^3$Mila - Quebec AI Institute \\
  $^4$Wuhan University 
  $^5$University of Montreal
  $^6$Tsinghua University
  $^7$MBZUAI \\
  \texttt{\{fanghua.ye.21, lijianjack\}@gmail.com} \\
}

\begin{document}
\maketitle
\footnotetext[1]{{ }Work done during an internship at Tencent Hunyuan.}
\footnotetext[2]{{ }$^{\ast}$ Equal Contribution.}
\footnotetext[3]{{ }$^{\dagger}$ Corresponding Author.}

\begin{abstract}

Large language models (LLMs) often produce content that contradicts or overlooks information provided in the input context, a phenomenon known as \textit{faithfulness hallucination}. In this paper, we propose \textbf{Context-Fidelity Boosting (CFB)}, a lightweight and general decoding-time framework that reduces such hallucinations by increasing the generation probability of source-supported tokens. Motivated by logit-shaping principles from watermarking techniques, CFB applies additive token-level logit adjustments based on a token's degree of support from the input context. Specifically, we develop three boosting strategies: \textit{static boosting}, which applies a fixed bias to source-supported tokens; \textit{context-aware boosting}, which scales this bias using the divergence between next-token distributions with and without context; and \textit{token-aware boosting}, which further redistributes the adaptive bias according to local relevance estimated from source-position attention and source-scoped semantic similarity. CFB requires no retraining or architectural changes, making it compatible with a wide range of LLMs. Experiments on summarization and question answering tasks across multiple open-source LLMs show that CFB consistently improves faithfulness metrics with minimal generation overhead. Our implementation is fully open-sourced.\footnote{\url{https://github.com/weixuzhang/CFB}}

\end{abstract}

\section{Introduction}

Large Language Models (LLMs) have demonstrated remarkable capabilities across a wide range of natural language tasks.
In many practical settings, however, models are expected to generate outputs that faithfully follow user-provided context, such as in retrieval-augmented generation (RAG), conversational information retrieval~\citep{adaptivepersonalizedcir}, summarization~\citep{laban2024summaryhaystackchallengelongcontext}, and question answering~\citep{chen2025ruleragruleguidedretrievalaugmentedgeneration}. When external evidence conflicts with a model's internal parametric memory, the generated content may contradict, ignore, or distort the provided context~\citep{mallen-etal-2023-trust,liu-etal-2024-untangle}, leading to \textbf{faithfulness hallucination}, outputs that are fluent and plausible but inconsistent with the input.

Faithfulness hallucination is distinct from the more commonly studied \textit{factuality hallucination}, which concerns incorrect or fabricated facts regardless of context. In high-stakes domains such as healthcare~\citep{zhu2024realmragdrivenenhancementmultimodal}, legal~\citep{cui2024chatlawmultiagentcollaborativelegal}, and finance~\citep{Lee_2025}, it is essential for model outputs to remain faithful to the given input even when the model's internal knowledge offers plausible but contextually irrelevant or conflicting alternatives.

\begin{figure}[t]
  \centering
  \includegraphics[width=\columnwidth]{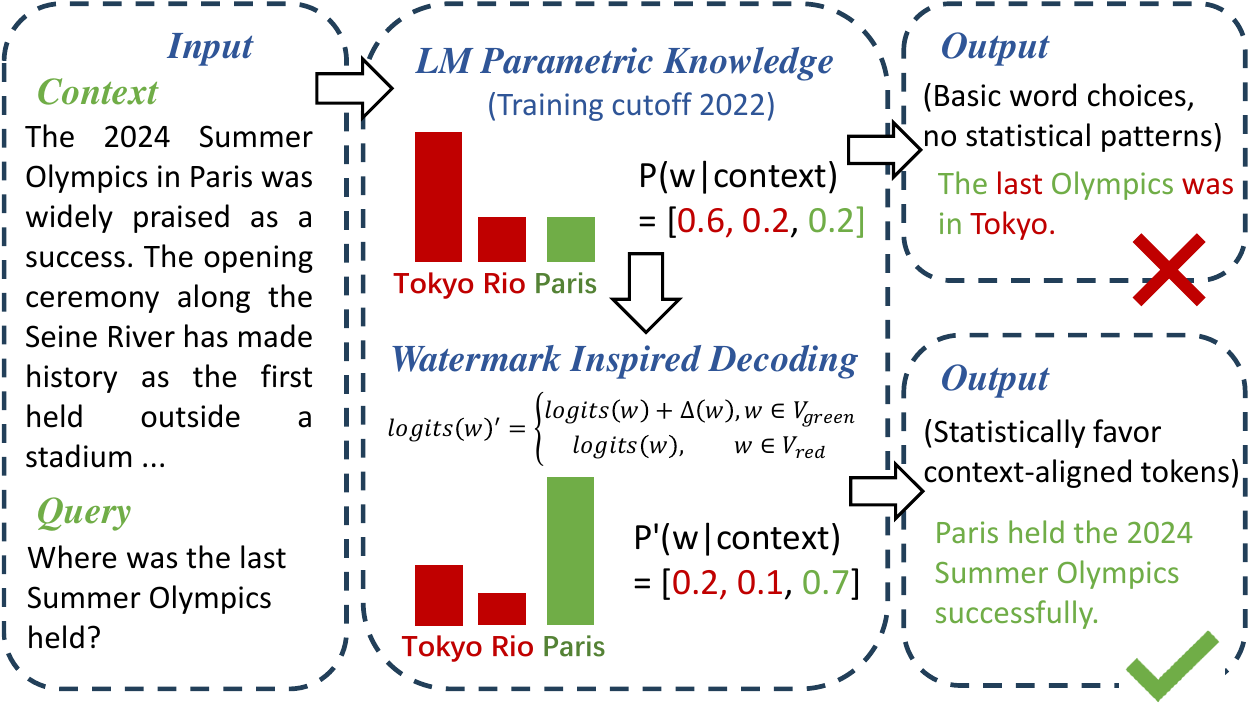}
  \caption{Illustration of context-faithful decoding: Traditional decoding relies on parametric knowledge (favoring ``Tokyo''), while our logit-shaping approach dynamically adjusts token probabilities to better align with the given context about ``Paris 2024''.}
  \label{fig:introduction}
  \vspace{-0.5cm}
\end{figure}

Existing methods for mitigating faithfulness hallucination generally fall into three categories: (1) training-time approaches that require fine-tuning or architectural changes~\citep{DBLP:journals/corr/abs-2406-11267}, (2) prompting techniques that rely on hand-crafted inputs and may behave inconsistently across tasks or models~\citep{DBLP:journals/corr/abs-2409-10790}, and (3) decoding-time interventions that modify generation behavior at inference time~\citep{DBLP:conf/naacl/ShiHLTZY24,DBLP:journals/corr/abs-2409-07394}. Among these, decoding-time methods are especially attractive because they are efficient, model-agnostic, and easy to deploy. However, many existing approaches still face a difficult trade-off between enforcing context fidelity and preserving fluency, or they depend on carefully tuned contrastive objectives and heuristic control rules.

In this work, we introduce \textbf{Context-Fidelity Boosting (CFB)}, a decoding-time framework that improves context alignment by selectively adjusting token probabilities during generation. CFB is inspired by logit-shaping mechanisms originally developed for text watermarking~\citep{kirchenbauer2024watermarklargelanguagemodels,liu2024semanticinvariantrobustwatermark,liu2024adaptivetextwatermarklarge}. While watermarking methods bias generation toward designated token sets in order to embed detectable signals, our goal is different: we use a similar logit-shaping principle to gently favor tokens supported by the input context, thereby reducing faithfulness hallucination without degrading fluency.

CFB operates through three levels of control.
\textit{Static boosting} applies a fixed bias to source-supported tokens.
\textit{Context-aware boosting} scales this bias using the divergence between context-aware and context-free next-token distributions, allowing the method to adapt to the degree of contextual influence.
\textit{Token-aware boosting} further redistributes the adaptive bias across source-supported tokens according to token-level relevance, combining source-position attention with source-scoped semantic similarity to provide finer-grained control during decoding.
Importantly, CFB requires no retraining or architectural modification and introduces only lightweight overhead during inference. This makes it a practical framework for improving faithfulness in real-world deployments.

Our key contributions are as follows:
\begin{itemize}[leftmargin=*]
    \item We propose CFB, a lightweight and model-agnostic decoding framework that improves contextual faithfulness while preserving output quality, making it especially suitable for high-stakes context-grounded generation.
    
    \item We develop a three-level boosting mechanism with static, context-aware, and token-aware variants, enabling finer-grained and more flexible control over context-sensitive decoding through additive logit shaping.
    
    \item We demonstrate the effectiveness of CFB across multiple model scales and tasks, including summarization and question answering, showing consistent gains in faithfulness metrics with minimal decoding overhead.
\end{itemize}

\section{Related Work}

\subsection{Faithfulness Hallucinations in LLMs}
Despite their impressive capabilities, LLMs frequently generate hallucinated content~\cite{DBLP:journals/corr/abs-2406-19354, DBLP:conf/emnlp/ChuangQHKKG24, DBLP:journals/corr/abs-2410-03727}. Recent studies distinguish two major forms of hallucination. \emph{Factuality hallucination}~\cite{DBLP:journals/corr/abs-2408-12325} arises when model outputs diverge from verifiable real-world facts, such as incorrect historical dates or fabricated attributions. In contrast, \emph{faithfulness hallucination}~\cite{DBLP:conf/nips/WuWZ24, DBLP:journals/corr/abs-2406-17519} occurs when generated content contradicts, ignores, or fabricates information relative to the provided input context, such as unsupported details in a summary. The latter is particularly problematic in context-grounded settings including summarization, retrieval-augmented generation, and question answering, where external evidence may conflict with a model's parametric knowledge acquired during pretraining. To evaluate faithfulness, prior work has proposed a range of metrics, including semantic similarity, entailment measures, and fact-checking frameworks~\cite{DBLP:conf/acl/NiuWZXSZS024, DBLP:journals/corr/abs-2404-05904}.

\subsection{Existing Mitigation Methods}

\begin{figure*}[t]
  \centering
  \includegraphics[width=0.9\linewidth]{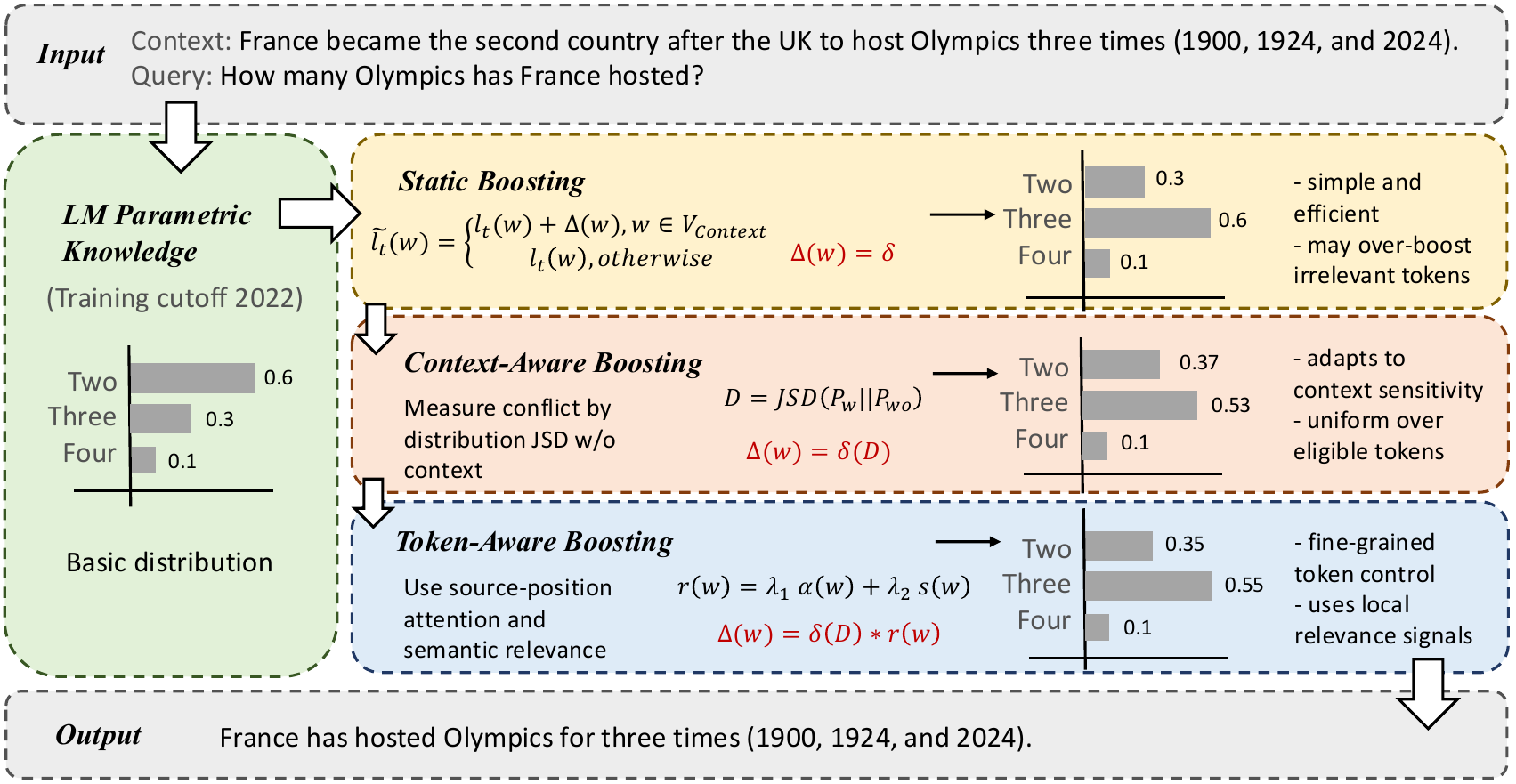}
  \caption{Overview of the proposed \textbf{Context-Fidelity Boosting (CFB)} framework. CFB applies additive logit shaping through three strategies of increasing adaptivity: (1) \textit{Static Boosting}, which uniformly increases logits of source-supported tokens by a fixed value; (2) \textit{Context-Aware Boosting}, which scales the boost using the divergence between context-aware and context-free next-token distributions; and (3) \textit{Token-Aware Boosting}, which further redistributes the adaptive boost according to token-level relevance estimated from attention over source positions and source-scoped semantic similarity.}
  \label{fig:framework}
  \vspace{-0.3cm}
\end{figure*}

Prior work has explored hallucination mitigation at different stages of the LLM pipeline~\cite{DBLP:journals/corr/abs-2311-05232}. Training-time approaches introduce architectural or objective modifications, such as enhanced attention or knowledge-aware training, but often require substantial computation and exhibit limited cross-domain generalization~\cite{DBLP:journals/corr/abs-2401-01313}. Prompting-based methods, including chain-of-thought reasoning~\citep{wei2023chainofthoughtpromptingelicitsreasoning} and self-consistency, provide model-agnostic alternatives but show variable effectiveness across models and tasks~\cite{DBLP:journals/corr/abs-2406-06950}. Decoding-time interventions directly modify generation behavior at inference time, for example through constrained decoding, contrastive decoding, or adaptive reweighting of token probabilities, yet frequently face a trade-off between enforcing faithfulness and preserving fluency~\cite{DBLP:journals/corr/abs-2410-18860,opendecoder}.

Among these approaches, decoding-time methods are particularly attractive because they do not require retraining and can be applied to a broad range of open-weight LLMs. Our method belongs to this family, but differs from prior approaches in two important ways. First, rather than contrasting full distributions or imposing hard decoding constraints, CFB uses lightweight additive logit shaping to favor tokens supported by the external context. Second, CFB provides multiple levels of control, ranging from a fixed context bias to sample-level adaptive scaling and token-aware redistribution, enabling a simple and interpretable trade-off between contextual fidelity and generation quality.

Related cache-based and memory-augmented decoding methods improve coherence by boosting tokens from recent generation history, primarily for better language modeling or long-range dependency modeling. Our token-aware variant is also related in spirit to relevance reweighting approaches that refine attention or ranking signals~\citep{reattn}. Unlike these methods, however, CFB operates directly at decoding time and explicitly boosts tokens grounded in the external input context, with the goal of improving contextual faithfulness rather than general language modeling quality.

\subsection{Watermarking in LLMs}

Recent work on text watermarking has demonstrated that subtle logit shaping can effectively steer model outputs while preserving generation quality. These methods typically partition the vocabulary into ``green'' and ``red'' token sets and adjust token probabilities to embed detectable statistical patterns~\cite{DBLP:conf/iclr/LiuPHM024}. Subsequent advances have proposed soft or adaptive watermarking schemes that dynamically adjust token probabilities based on context~\citep{kirchenbauer2024watermarklargelanguagemodels}, semantic invariance constraints~\citep{liu2024semanticinvariantrobustwatermark}, and theoretical analyses of the trade-off between watermark strength and naturalness~\citep{golowich2024editdistancerobustwatermarks}. While watermarking aims to embed identifiable signals for downstream detection, our work repurposes similar logit-shaping mechanisms to improve context faithfulness. Instead of favoring a predefined statistical token subset, CFB uses additive logit shaping to bias generation toward tokens supported by the input context, with adaptive scaling and token-aware control. In this sense, CFB transfers the controllability insight of watermarking to the problem of reducing faithfulness hallucination in context-grounded generation.

\section{Methodology}

\begin{table*}[t]
\small
\begin{center}
\begin{tabular}{@{}p{0.97\textwidth}@{}}
\toprule
\textbf{Algorithm 1: Context-Fidelity Boosting (CFB)} \\[0.2em]
\midrule
\begin{tabular}{@{}p{0.15\textwidth}p{0.8\textwidth}@{}}
\textbf{Input:} & Context passage $C$, query $Q$, language model $M$ \\
\textbf{Parameters:} & $\delta$, $\delta_{\min}$, $\delta_{\max}$, $\lambda_1$, $\lambda_2$ \\
\textbf{Output:} & Generated output with boosted context fidelity
\end{tabular} \\[0.4em]
\midrule
\\[-0.8em]
\begin{tabular}{@{}p{0.08\textwidth}p{0.87\textwidth}@{}}
1: & $S \leftarrow$ ResolveSourceSpan$(C, Q)$ \hfill // source-supported span \\
2: & $T_S \leftarrow$ UniqueTokens$(S)$,\quad $\text{pos}(w) \leftarrow$ Occurrences$(w, S)$,\quad $s(w) \leftarrow$ SourceScopedSemanticSimilarity$(w, S)$ \\
3: & $D \leftarrow \mathrm{JSD}(M(C+Q), M(Q))$,\quad $\delta_D \leftarrow \delta_{\min} + (\delta_{\max}-\delta_{\min})\cdot D$ \\
4: & $output\_ids \leftarrow \text{Tokenize}(C+Q)$ \\
5: & \textbf{while} not terminated \textbf{do} \\
6: & \quad $l_t \leftarrow M(output\_ids)[-1]$ \\
7: & \quad \textbf{if} mode is ``static'' \textbf{then \ } $\tilde{l}_t(w) \leftarrow l_t(w)+\delta,\ \forall w \in T_S$ \\
8: & \quad \textbf{else if} mode is ``context-aware'' \textbf{then \ } $\tilde{l}_t(w) \leftarrow l_t(w)+\delta_D,\ \forall w \in T_S$ \\
9: & \quad \textbf{else if} mode is ``token-aware'' \textbf{then} \\
10: & \qquad $a_t(p) \leftarrow$ GetAttentionScores over source positions in $S$ \\
11: & \qquad $\alpha_t(w) \leftarrow \mathrm{Agg}(\{a_t(p): p \in \text{pos}(w)\}),\quad r_t(w) \leftarrow \lambda_1 \alpha_t(w)+\lambda_2 s(w)$ \\
12: & \qquad $\hat{r}_t(w) \leftarrow$ Normalize$(r_t(w))$ over $w \in T_S$,\quad $\tilde{l}_t(w) \leftarrow l_t(w)+\delta_D \hat{r}_t(w)$ \\
13: & \quad $P^* \leftarrow \mathrm{Softmax}(\tilde{l}_t)$,\quad $next\_token \leftarrow \mathrm{Sample}(P^*)$ \\
14: & \quad $output\_ids \leftarrow [output\_ids; next\_token]$ \\
15: & \textbf{return} Decode$(output\_ids)$
\end{tabular} \\
\bottomrule
\end{tabular}
\end{center}
\vspace{-0.2cm}
\caption{Pseudocode for Context-Fidelity Boosting (CFB). The token-aware variant first computes a sample-level adaptive boost from the divergence between next-token distributions with and without context, and then redistributes this boost across source-supported tokens using attention-based and semantic relevance signals.}
\label{alg:cfb}
\end{table*}

We introduce \textbf{Context-Fidelity Boosting (CFB)}, a decoding-time framework designed to reduce faithfulness hallucination by adaptively adjusting token probabilities based on their support from the input context. Motivated by recent logit-shaping techniques used in text watermarking, CFB selectively promotes source-supported tokens during generation to better align outputs with the provided evidence, as illustrated in Figure~\ref{fig:framework}.

\subsection{Problem Formulation}

Given a context passage $C$ and a query $Q$, our goal is to enhance contextual fidelity by increasing the probability of generating tokens supported by the source content in $C$. Let $P(y_t \mid y_{<t}, C, Q)$ denote the generation probability at time step $t$, and let $V_S \subseteq V$ denote the set of vocabulary tokens appearing in the source span extracted from the input context. We reshape the logits as follows:
\vspace{-0.2cm}
\begin{equation}
\tilde{l}_t(w) =
\begin{cases}
l_t(w) + \Delta_t(w), & \text{if } w \in V_S,\\
l_t(w), & \text{otherwise}.
\end{cases}
\end{equation}

Here, $l_t(w)$ is the original logit for token $w$, and $\Delta_t(w)$ is a boosting factor determined by the support that token receives from the input context.

\subsection{Context-Fidelity Boosting Framework}

We propose a three-level logit-shaping strategy, progressing from fixed boosting to sample-level adaptive scaling and token-aware redistribution.

\subsubsection{Static Boosting}
The simplest strategy assigns a fixed boost $\delta$ to all tokens in the source-supported vocabulary:
\begin{equation}
\Delta_t(w) = \delta.
\end{equation}

While effective in encouraging context preference, this strategy does not account for variation across inputs or for differences in the relevance and importance of individual source-supported tokens.

\subsubsection{Context-Aware Boosting}

To dynamically adjust the boost according to the influence of the context, we compute the Jensen-Shannon divergence between context-aware and context-free next-token distributions:
\vspace{-0.2cm}
\begin{equation}
D = \mathrm{JSD}(P_{\mathrm{w}} \| P_{\mathrm{wo}}),
\end{equation}
where $P_{\mathrm{w}}$ and $P_{\mathrm{wo}}$ denote the predicted next-token distributions with and without context, respectively. $\mathrm{JSD}(\cdot)$ is bounded in $[0,1]$ under base-2 logarithms.

We use $D$ to scale the boost adaptively:
\begin{equation}
\Delta_t(w) = \delta(D) = \delta_{\min} + (\delta_{\max} - \delta_{\min}) \cdot D,
\end{equation}
where $\delta_{\min}$ and $\delta_{\max}$ define the boosting range. Intuitively, larger divergence indicates that the context substantially changes the model's next-token preference, and therefore warrants a stronger context-fidelity bias.

\subsubsection{Token-Aware Boosting}

The most fine-grained strategy retains the sample-level adaptive boost and further redistributes it across source-supported tokens according to token-level relevance. Specifically, at decoding step $t$, we first compute attention scores over positions in the extracted source span. For each token $w \in V_S$, we aggregate the attention mass over all of its occurrences in the source:
\begin{equation}
\alpha_t(w) = \mathrm{Agg}\left(\{a_t(p) : p \in \mathcal{P}(w, C)\}\right),
\end{equation}
where $\mathcal{P}(w, C)$ denotes the set of source positions containing token $w$, and $\mathrm{Agg}(\cdot)$ is an aggregation operator. In our implementation, we use summation to accumulate attention over repeated occurrences.

We further compute a source-scoped semantic relevance score by averaging the cosine similarity between the embedding of token $w$ and the embeddings of tokens in the extracted source span:
\begin{equation}
s(w) = \frac{1}{|S|} \sum_{c \in S} \mathrm{cosine}(e_w, e_c),
\end{equation}
where $S$ denotes the extracted source span, and $e_w$ and $e_c$ are token embeddings. The token relevance score is then defined as
\begin{equation}
r_t(w) = \lambda_1 \alpha_t(w) + \lambda_2 s(w),
\end{equation}
with $\lambda_1 + \lambda_2 = 1$.

To maintain a comparable overall boost scale across eligible tokens, we normalize the relevance scores:
\begin{equation}
\hat{r}_t(w) = \frac{r_t(w)}{\frac{1}{|V_S|}\sum_{u \in V_S} r_t(u)}.
\end{equation}
The final token-aware boost is then
\begin{equation}
\Delta_t(w) = \delta(D) \cdot \hat{r}_t(w).
\end{equation}

This design enables finer-grained control than context-aware boosting by allocating larger boosts to source-supported tokens that are estimated to be more relevant under the current decoding state. In all experiments, we set $\lambda_1 = 0.6$ and $\lambda_2 = 0.4$.

\subsection{Implementation Details}
Table~\ref{alg:cfb} outlines the full CFB decoding algorithm. In practice, we first resolve the source span from the input prompt and restrict boosting to tokens supported by this span, which reduces interference from prompt scaffolding and instruction text. For token-aware boosting, source-scoped semantic relevance is precomputed once per example, while attention over source positions is recomputed at each decoding step to capture the current decoding state. Token-level relevance is then obtained by aggregating attention over token occurrences and normalizing the resulting scores before applying the final additive logit bias. This design keeps the computation lightweight while preserving fine-grained control during generation.

\begin{table*}[!ht]
\small
\centering
\resizebox{\textwidth}{!}{
\begin{tabular}{llcccccc}
\toprule
\multirow{2}{*}{\textbf{Model}} & \multirow{2}{*}{\textbf{Method}} & \multicolumn{3}{c}{\textbf{CNN/DM}} & \multicolumn{3}{c}{\textbf{XSum}} \\
\cmidrule(lr){3-5} \cmidrule(lr){6-8}
& & \textbf{ROUGE-L} & \textbf{FactKB} & \textbf{BERT-P} & \textbf{ROUGE-L} & \textbf{FactKB} & \textbf{BERT-P} \\
\midrule
\multirow{6}{*}{Mistral-7B}
& CAD \citep{DBLP:conf/naacl/ShiHLTZY24} & 33.19 & 96.39 & \textbf{89.65} & \textbf{16.57} & 39.22 & 89.93 \\
& ADACAD \citep{DBLP:journals/corr/abs-2409-07394} & 25.71 & 85.67 & 86.20 & 14.46 & 29.19 & 86.42 \\
& COIECD \cite{DBLP:conf/acl/YuanYWLZL24} & 22.65 & 75.08 & 84.84 & 11.93 & 27.09 & 84.27 \\
\cmidrule(r){2-8}
& Static CFB (ours) & 34.22 & 95.37 & 89.48 & 14.66 & 54.25 & 89.55 \\
& Context-aware CFB (ours)& 34.05 & 95.93 & 89.51 & 14.56 & \textbf{56.02} & \textbf{89.64} \\
& Token-aware CFB (ours) & \textbf{34.52} & \textbf{96.87} & 89.44 & 14.75 & 48.69 & 88.85 \\
\midrule
\multirow{6}{*}{Llama2-13B}
& CAD \citep{DBLP:conf/naacl/ShiHLTZY24} & 35.63 & 97.26 & 89.38 & 13.96 & 26.91 & 88.86 \\
& ADACAD \citep{DBLP:journals/corr/abs-2409-07394} & 24.10 & 89.97 & 85.60 & 10.74 & 38.83 & 83.68 \\
& COIECD \cite{DBLP:conf/acl/YuanYWLZL24} & 19.37 & 80.19 & 83.57 & 9.49 & 9.51 & 84.16\\
\cmidrule(r){2-8}
& Static CFB (ours) & 37.40 & \textbf{98.85} & 89.61 & 13.77 & 54.38 & 89.53 \\
& Context-aware CFB (ours)& \textbf{37.52} & 98.69 & 89.62 & 14.62 & \textbf{55.02} & 89.49 \\
& Token-aware CFB (ours) & 36.16 & 97.24 & \textbf{89.83} & \textbf{15.25} & 37.91 & \textbf{89.57} \\
\midrule
\multirow{6}{*}{Llama3-8B}
& CAD \citep{DBLP:conf/naacl/ShiHLTZY24} & 35.92 & 94.57 & 89.07 & 12.92 & 45.77 & 87.05 \\
& ADACAD \citep{DBLP:journals/corr/abs-2409-07394} & 21.80 & 87.67 & 84.60 & 8.69 & 42.81 & 82.07 \\
& COIECD \cite{DBLP:conf/acl/YuanYWLZL24} & 19.11 & 83.46 & 83.96 & 10.59 & 51.90 & 83.80 \\
\cmidrule(r){2-8}
& Static CFB (ours) & \textbf{36.79} & 95.15 & 89.63 & 12.46 & 66.33 & \textbf{88.69} \\
& Context-aware CFB (ours) & 36.78 & \textbf{97.23} & \textbf{89.85} & 12.59 & \textbf{66.85} & 88.67 \\
& Token-aware CFB (ours) & 35.81 & 94.31 & 89.38 & \textbf{13.23} & 55.29 & 88.45\\

\bottomrule
\end{tabular}}
\caption{Main summarization results on CNN/DM and XSum. We report ROUGE-L, FactKB, and BERT-P. Best results for each model are shown in \textbf{bold}.}
\label{tab:sum_results}
\end{table*}

\begin{table*}[!ht]
\small
\centering
\resizebox{\textwidth}{!}{
\begin{tabular}{llcccccccc}
\toprule
\multirow{2}{*}{\textbf{Model}} & \multirow{2}{*}{\textbf{Method}} & \multicolumn{4}{c}{\textbf{NQ-Synth}} & \multicolumn{4}{c}{\textbf{NQ-Swap}} \\
\cmidrule(lr){3-6} \cmidrule(lr){7-10}
& & \textbf{ROUGE-L} & \textbf{FactKB} & \textbf{BERT-P} & \textbf{Acc} & \textbf{ROUGE-L} & \textbf{FactKB} & \textbf{BERT-P} & \textbf{Acc} \\
\midrule
\multirow{6}{*}{Mistral-7B}
& CAD \citep{DBLP:conf/naacl/ShiHLTZY24} & 26.64 & 57.92 & 87.07 & 48.20 & 35.11 & 49.46 & 74.56 & 50.06 \\
& ADACAD \citep{DBLP:journals/corr/abs-2409-07394} & 7.71 & \textbf{67.46} & 86.78 & 48.30 & \textbf{36.41} & \textbf{67.83} & 87.53 & \textbf{73.79} \\
& COIECD \cite{DBLP:conf/acl/YuanYWLZL24} & 12.20 & 48.41 & 85.65 & 20.70 & 5.01 & 27.07 & 83.60 & 3.15 \\
\cmidrule(r){2-10}
& Static CFB (ours) & 24.72 & 54.74 & 87.73 & 56.60 & 16.29 & 51.91 & 88.36 & 36.08 \\
& Context-aware CFB (ours) & 24.67 & 54.74 & 87.73 & 56.50 & 16.26 & 51.94 & 88.37 & 36.01 \\
& Token-aware CFB (ours) & \textbf{28.77} & 50.21 & \textbf{88.18} & \textbf{60.10} & 17.36 & 48.46 & \textbf{88.50} & 35.73 \\
\midrule
\multirow{6}{*}{Llama2-13B}
& CAD \citep{DBLP:conf/naacl/ShiHLTZY24} & 29.34 & 56.45 & 86.78 & 47.90 & \textbf{23.22} & 35.59 & 84.95 & \textbf{44.91} \\
& ADACAD \citep{DBLP:journals/corr/abs-2409-07394} & 12.68 & \textbf{65.20} & 86.52 & 39.70 & 20.11 & 34.60 & 82.25 & 74.21 \\
& COIECD \cite{DBLP:conf/acl/YuanYWLZL24} & 14.67 & 43.13 & 85.02 & 20.60 & 2.52 & 24.24 & 74.13 & 1.50 \\
\cmidrule(r){2-10}
& Static CFB (ours) & 34.43 & 48.77 & 83.70 & 56.00 & 12.66 & 55.27 & 88.32 & 26.03 \\
& Context-aware CFB (ours)& 34.43 & 48.95 & 83.71 & 56.00 & 12.64 & \textbf{55.35} & \textbf{88.33} & 26.03 \\
& Token-aware CFB (ours) & \textbf{39.05} & 49.68 & \textbf{87.64} & \textbf{64.00} & 13.10 & 54.19 & 81.20 & 11.13 \\
\midrule
\multirow{6}{*}{Llama3-8B}
& CAD \citep{DBLP:conf/naacl/ShiHLTZY24} & 28.19 & 32.26 & 86.50 & 66.80 & \textbf{26.40} & 33.66 & 87.05 & 58.49 \\
& ADACAD \citep{DBLP:journals/corr/abs-2409-07394} & 6.26 & \textbf{51.36} & 86.50 & 48.40 & 12.52 & 39.14 & 85.82 & \textbf{86.50} \\
& COIECD \cite{DBLP:conf/acl/YuanYWLZL24} & 15.24 & 21.97 & 84.54 & 32.10 & 5.73 & 26.15 & 84.49 & 6.33 \\
\cmidrule(r){2-10}
& Static CFB (ours) & 29.87 & 44.53 & \textbf{88.34} & 73.10 & 14.14 & \textbf{48.95} & 88.48 & 34.88 \\
& Context-aware CFB (ours)& 29.87 & 44.50 & 88.33 & 73.10 & 14.16 & 48.93 & \textbf{88.49} & 34.91 \\
& Token-aware CFB (ours) & \textbf{32.90} & 45.94 & 88.13 & \textbf{73.40} & 14.54 & 40.92 & 87.99 & 32.43 \\
\bottomrule
\end{tabular}}
\caption{Main QA-style generation results on NQ-Synth and NQ-Swap. We report ROUGE-L, FactKB, BERT-P, and accuracy (\%). Best results for each model are shown in \textbf{bold}.}
\label{tab:qa_results}
\end{table*}

\section{Experiments}

\subsection{Experiment Setup}
\paragraph{Models} We evaluate our method on several state-of-the-art LLMs, including Llama2-13B-chat-hf, Llama3-8B-Instruct, and Mistral-7B-Instruct.

\paragraph{Datasets} We consider two types of tasks.
\begin{itemize}[leftmargin=*]
    \item \textbf{Summarization:} We use CNN-DM \citep{see-etal-2017-get} and XSum \citep{narayan-etal-2018-dont} to evaluate the model's ability to generate faithful and context-grounded summaries. For XSum, we randomly sample 500 examples for evaluation. For this task, we measure ROUGE-L \citep{lin-2004-rouge} for summary quality, factKB \citep{feng2023factkbgeneralizablefactualityevaluation} for knowledge consistency, and BERT-P \citep{zhang2020bertscoreevaluatingtextgeneration} for semantic preservation.
    \item \textbf{Question Answering:} We use NQ-SWAP \citep{longpre-etal-2021-entity} and NQ-Synth \citep{DBLP:journals/corr/abs-2409-07394} to evaluate the model's ability to leverage context information. NQ-SWAP contains synthetic knowledge conflicts, while NQ-Synth consists of examples where context aligns with the model's parametric knowledge. For this task, we also report accuracy.
\end{itemize}

\paragraph{Baselines} We compare our method against several strong decoding-time baselines: Context-Aware Decoding (CAD) \citep{DBLP:conf/naacl/ShiHLTZY24}, which uses a fixed hyperparameter to control adjustment of output probabilities; Adaptive Context-Aware Decoding (ADACAD) \citep{DBLP:journals/corr/abs-2409-07394}, which dynamically infers adjustment based on Jensen-Shannon divergence; and Contextual Information-Entropy Constraint Decoding (COIECD) \cite{DBLP:conf/acl/YuanYWLZL24}, which employs distinct strategies for conflicting and non-conflicting tokens. For consistent comparison, we use top-$p$ sampling across all methods under a zero-shot setting, with hyperparameters following their original papers.

\subsection{Results}

\paragraph{Overall Performance}
Overall, CFB remains competitive with strong decoding-time baselines across both summarization and QA tasks. A clear pattern is that CFB is particularly effective when the task benefits from reinforcing source-supported content, such as summarization and the complementary-knowledge setting in NQ-Synth. In contrast, on NQ-Swap, where context explicitly conflicts with parametric knowledge, stronger contrastive suppression methods such as ADACAD often remain more effective.

\paragraph{Summarization Performance}

For summarization tasks, as shown in Table~\ref{tab:sum_results}, CFB consistently improves over strong baselines on CNN/DM across all three models. On Mistral-7B, Token-aware CFB achieves the best ROUGE-L (34.52) and FactKB (96.87), while CAD remains slightly better on BERT-P (89.65). On Llama2-13B, Context-aware CFB achieves the best ROUGE-L (37.52), Static CFB attains the highest FactKB (98.85), and Token-aware CFB obtains the best BERT-P (89.83). On Llama3-8B, Static CFB gives the best ROUGE-L (36.79), whereas Context-aware CFB achieves the strongest factual consistency and semantic preservation, with FactKB 97.23 and BERT-P 89.85.

\begin{table}[!t]
\small
\centering
\resizebox{\columnwidth}{!}{
\begin{tabular}{lcccccc}
\toprule
& \multicolumn{3}{c}{\textbf{Human Ratings}} & \multicolumn{3}{c}{\textbf{LLM Evaluation}} \\
\cmidrule(lr){2-4} \cmidrule(lr){5-7}
\textbf{Method} & \textbf{Faith.} & \textbf{Flu.} & \textbf{Info.} & \textbf{Consist.} & \textbf{Hall.} & \textbf{Contra.} \\
\midrule
CAD & 3.82 & 4.15 & 3.76 & 0.83 & 1.24 & 0.12 \\
ADACAD & 4.03 & 4.21 & 3.89 & 0.87 & 0.95 & 0.09 \\
Token-aware CFB (Ours) & \textbf{4.31} & 4.18 & \textbf{4.12} & \textbf{0.91} & \textbf{0.67} & \textbf{0.05} \\
\bottomrule
\end{tabular}}
\caption{Human and LLM-based evaluation results. Faith., Flu., and Info. denote faithfulness, fluency, and informativeness, respectively. Consist. denotes consistency, Hall. denotes the average number of hallucinations per output, and Contra. denotes contradiction rate. Human ratings are on a 1--5 scale, where higher is better.}
\label{tab:human_eval}
\vspace{-0.5cm}
\end{table}

On XSum, the pattern is more model-dependent and highlights a clearer trade-off across evaluation metrics. For Mistral-7B, CAD still achieves the best ROUGE-L (16.57), but Context-aware CFB substantially improves factual consistency and semantic preservation, reaching the best FactKB (56.02) and BERT-P (89.64). For Llama2-13B, all three CFB variants outperform baselines by large margins, with Token-aware CFB achieving the best ROUGE-L (15.25), Context-aware CFB the best FactKB (55.02), and Token-aware CFB the best BERT-P (89.57). For Llama3-8B, Token-aware CFB achieves the best ROUGE-L (13.23), while Context-aware CFB achieves the best FactKB (66.85) and Static CFB the best BERT-P (88.69). Overall, these results suggest that on XSum, CFB reliably strengthens faithfulness-related metrics, while the strongest variant depends on both the underlying model and the desired trade-off between lexical overlap, factual consistency, and semantic preservation.

\paragraph{Question Answering Performance}
In QA tasks, shown in Table~\ref{tab:qa_results}, we observe a clear difference between NQ-Synth and NQ-Swap. On NQ-Synth, where the provided context is complementary to the model's parametric knowledge, CFB consistently improves generation quality and answer accuracy. Token-aware CFB achieves the best accuracy for all three models: 60.10 on Mistral-7B, 64.00 on Llama2-13B, and 73.40 on Llama3-8B. It also delivers the best ROUGE-L for all three models, indicating that token-level redistribution is particularly helpful when the model can benefit from reinforcing context-supported evidence.

On NQ-Swap, however, the trend differs. ADACAD performs best on accuracy for all three models, achieving 73.79 on Mistral-7B, 74.21 on Llama2-13B, and 86.50 on Llama3-8B. In comparison, CFB variants generally lag behind on this conflict-heavy setting, although they remain competitive on some faithfulness-related metrics. For example, Context-aware CFB achieves the best FactKB (55.35) and BERT-P (88.33) on Llama2-13B, and Static/Context-aware CFB achieve the strongest FactKB and BERT-P among CFB variants on Llama3-8B. These results suggest that CFB is especially effective when context should be amplified, whereas methods explicitly designed to suppress parametric priors remain stronger when context and internal knowledge are in direct conflict.

\paragraph{Model-Specific Analysis}
CFB's behavior also varies across model architectures. On CNN/DM, all three models benefit substantially from CFB, but the strongest variant differs: Token-aware CFB works best for Mistral-7B, Context-aware CFB is strongest on Llama2-13B, and Static or Context-aware CFB are most effective on Llama3-8B depending on the metric. On XSum, CFB provides especially large factuality gains over baselines for Llama2-13B and Llama3-8B, while Mistral-7B remains more competitive with CAD on ROUGE-L. In QA, Llama3-8B shows the largest gains from CFB on NQ-Synth, reaching 73.40 accuracy with Token-aware CFB, but also exhibits the sharpest gap to ADACAD on NQ-Swap. Taken together, these results suggest that CFB is robust for source-grounded generation, especially in summarization and complementary-knowledge QA, while its effectiveness under explicit knowledge conflict depends more strongly on model architecture and the choice of decoding strategy.

\subsection{Human Evaluation}

To complement automatic metrics, we further conduct human and LLM-based evaluation to assess generation quality from a qualitative perspective. We randomly sample 100 examples each from CNN-DM and NQ-Swap, and compare outputs from CAD, ADACAD, and Token-aware CFB. The results are reported in Table~\ref{tab:human_eval}.

\paragraph{Evaluation Protocol}
Three expert annotators independently rate each output on three dimensions: \textit{faithfulness}, \textit{fluency}, and \textit{informativeness}, using a 1--5 scale. For \textit{faithfulness}, a score of 1 indicates that the output contains major contradictions or unsupported content, while a score of 5 indicates that it is fully consistent with the input context. For \textit{fluency}, a score of 1 indicates severe grammatical or coherence issues, while a score of 5 indicates natural and fluent language. For \textit{informativeness}, a score of 1 indicates that the output is largely incomplete or irrelevant, while a score of 5 indicates that it is highly relevant and sufficiently informative.

\begin{table*}[t]
\small
\centering
\resizebox{0.97\textwidth}{!}{
\begin{tabular}{l|ccccccc}
\toprule
\textbf{Models} & \textbf{Base Model} & \textbf{CAD} & \textbf{ADACAD} & \textbf{COIECD} & \textbf{Static CFB} & \textbf{Context-aware CFB} & \textbf{Token-aware CFB}  \\
\midrule
FLOPS & 3.40e+12 & 4.92e+07 & 1.15e+08 & 1.31e+08 & 8.19e+07 & 9.83e+07 & 2.86e+08  \\
\bottomrule
\end{tabular}}
\caption{Estimated FLOPS per decoding step. ``Base Model'' reports standard transformer decoding cost, while other entries denote additional method-specific overhead.}

\label{tab:flops}
\vspace{-10pt}
\end{table*}

\paragraph{Human Evaluation Results}
As shown in Table~\ref{tab:human_eval}, Token-aware CFB achieves the best human rating on faithfulness (4.31) and informativeness (4.12), outperforming both CAD and ADACAD. Its fluency score (4.18) remains comparable to the baselines, indicating that stronger contextual grounding does not noticeably harm language quality. These results suggest that CFB improves factual alignment and content coverage while preserving readability.

\paragraph{LLM-based Evaluation}
We further use GPT-4o as an automatic judge with a structured rubric to evaluate consistency, hallucination, and contradiction. Token-aware CFB again performs best, achieving the highest consistency (0.91), the lowest hallucination count (0.67), and the lowest contradiction rate (0.05). This trend is consistent with the human judgments and provides additional evidence that CFB improves contextual reliability.

Overall, the human and LLM-based evaluations are consistent: Token-aware CFB improves faithfulness and informativeness without noticeably degrading fluency. This provides qualitative support for the gains observed in the automatic metrics.

\begin{table}[!t]
\small
\centering
\resizebox{\columnwidth}{!}{
\begin{tabular}{lccc}
\toprule
\textbf{Method Variant} & \textbf{ROUGE-L} & \textbf{FactKB} & \textbf{BERT-P} \\
\midrule
Token-aware CFB & \textbf{35.81} & \textbf{94.31} & \textbf{89.38} \\
- w/o attention & 35.60 & 93.74 & 88.48 \\
- w/o semantic & 4.45 & 66.84 & 67.68 \\
- w/o JSD & 35.24 & 93.60 & 88.43 \\
\bottomrule
\end{tabular}}
\caption{Ablation study of Token-aware CFB on Llama3-8B on CNN-DM. We report ROUGE-L, FactKB, and BERT-P.}
\label{tab:ablation}
\vspace{-0.3cm}
\end{table}

\subsection{Computational Efficiency}

We compute the estimated FLOPS per decoding step for each method by breaking down their component operations (e.g., attention, similarity, logit shaping). Calculations assume a standard Llama-like setup: batch size 1, sequence length 128, hidden size 4096, 32 layers, and context length 512.

As shown in Table~\ref{tab:flops}, all CFB variants are efficient relative to the base model. In particular, the Static and Context-aware CFB variants require less than $0.003\%$ of the base model’s FLOPS, while Token-aware CFB, though more compute-intensive, remains lightweight and offers finer control. Compared to baselines like CAD, ADACAD, and COIECD, our variants achieve favorable trade-offs between complexity and fidelity control.
We further report empirical runtime per decoding step in Table \ref{tab:efficiency} in Appendix \ref{app:efficiency}. The results confirm that CFB decoding is practical in real-world settings, with Static and Context-aware variants offering strong efficiency-performance balance.

\colorlet{Mycolor1}{green!5!orange!95!}
\definecolor{Mycolor2}{HTML}{5e1bea}
\definecolor{Mycolor3}{HTML}{ec198f} 
\begin{table*}[t]
\small
\centering
\resizebox{\textwidth}{!}{
\begin{tabular}{p{0.15\textwidth}|p{0.85\textwidth}}
\toprule
\multicolumn{1}{l}{\textbf{Component}} & \multicolumn{1}{l}{\textbf{Content}} \\
\midrule
Context & The Great Wall of China's actual length was determined in 2012 to be \textcolor{Mycolor1}{21,196 kilometers (13,171 miles)}, much longer than the previously believed \textcolor{Mycolor1}{8,850 kilometers}. This finding came after a detailed archaeological survey using advanced mapping techniques, which also revealed that the wall contained sections from more diverse historical periods than originally thought... \\
\midrule
Question & What is the actual length of the Great Wall of China according to the 2012 survey, and how does it compare to previous estimates? \\
\midrule
Gold Answer & According to the 2012 survey, the Great Wall is \textcolor{Mycolor3}{21,196 kilometers} long, which is much longer than the previous estimate of \textcolor{Mycolor3}{8,850 kilometers}. \\
\midrule
Standard (Greedy) & The Great Wall of China is approximately \textcolor{Mycolor2}{8,850 kilometers} long. \\
CAD & The 2012 survey found the Great Wall to be \textcolor{Mycolor2}{21,196 kilometers}, longer than previous estimates. \\
COIECD & The Great Wall is about \textcolor{Mycolor2}{13,171 miles (8,850 kilometers)} long based on traditional measurements. \\
CFB (Ours)& According to the 2012 survey, the Great Wall is \textcolor{Mycolor3}{21,196 kilometers} long, much longer than the previously believed \textcolor{Mycolor3}{8,850 kilometers}. \\
\bottomrule
\end{tabular}}
\caption{Qualitative comparison showing how different methods handle knowledge conflicts. }
\label{tab:case_study}
\end{table*}

\begin{figure*}[!ht]
\centering
\begin{subfigure}[b]{0.48\textwidth}
    \includegraphics[width=\textwidth]{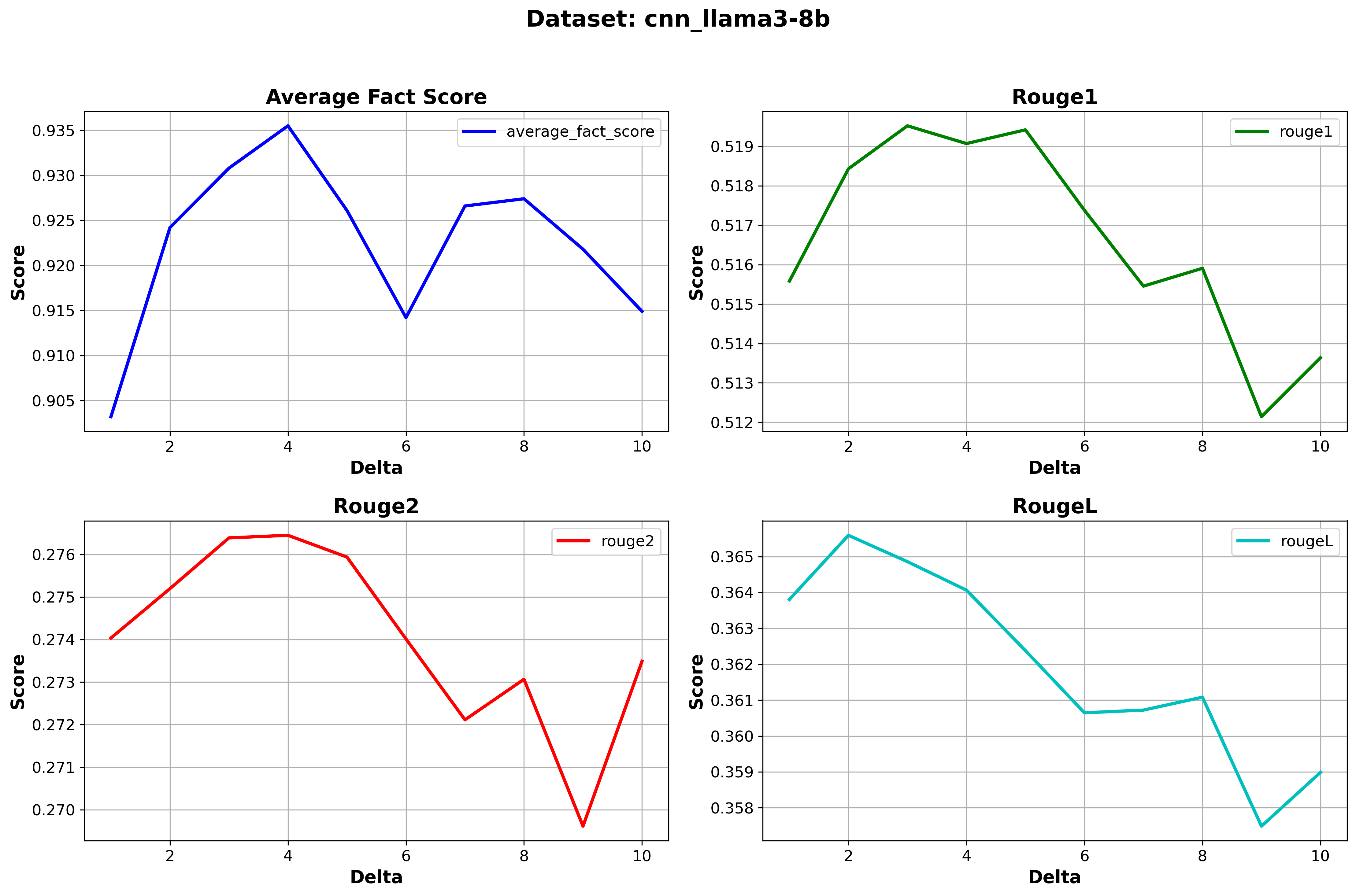}
    \caption{CNN-DM}
\end{subfigure}
\begin{subfigure}[b]{0.48\textwidth}
    \includegraphics[width=\textwidth]{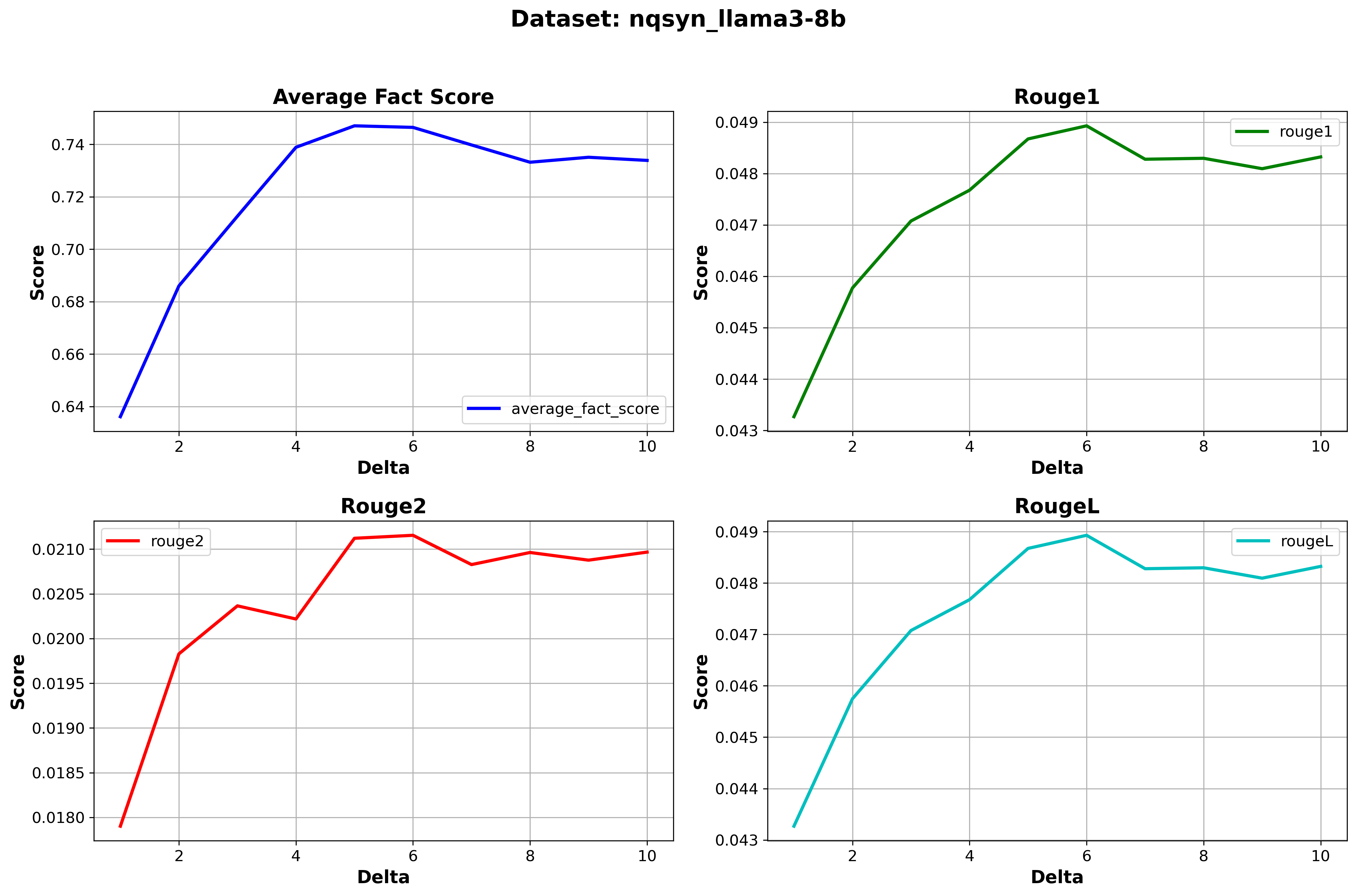}
    \caption{NQ-Synth} 
\end{subfigure}
\vspace{-0.1cm}
\caption{Impact of boost values ($\delta$) on fact scores and ROUGE metrics using Llama3-8B. We show the average fact score (top-left), ROUGE-1 (top-right), ROUGE-2 (bottom-left), and ROUGE-L (bottom-right) scores.}
\label{fig:boost_impact}
\vspace{-0.4cm}
\end{figure*}

\subsection{Ablation Study and Parameter Analysis}

To understand the impact of individual components in Token-aware CFB, we conduct ablation studies and parameter sensitivity analysis using Llama3-8B on CNN-DM.

\paragraph{Component Ablation}
We ablate the main components of Token-aware CFB in Table~\ref{tab:ablation}. The full model achieves the best performance on all three metrics, reaching 35.81 ROUGE-L, 94.31 FactKB, and 89.38 BERT-P. Removing attention causes only a modest drop across metrics, suggesting that the attention signal provides useful but limited additional benefit in the current formulation. Removing JSD also leads to consistent degradation, confirming that sample-level adaptive scaling remains important even when token-level reweighting is applied. In contrast, removing semantic similarity results in a severe collapse across all metrics, indicating that the semantic component plays a crucial stabilizing role in the token-aware variant. Taken together, these results suggest that Token-aware CFB benefits from combining global adaptive scaling with local relevance signals, while semantic relevance is currently the most critical component for maintaining stable and effective behavior.

\paragraph{Parameter Analysis}
We further analyze the effect of the boost value $\delta$ in Figure~\ref{fig:boost_impact}. For CNN-DM, moderate boost values yield the best trade-off between faithfulness and generation quality, while overly large boosts lead to degraded performance. On NQ-Synth, performance is more stable across a wider range of boost values, suggesting that stronger context reinforcement is more tolerable when the input context complements the model's parametric knowledge.

\subsection{Case Studies}

\paragraph{Case 1: High Knowledge Conflict}
As shown in Table~\ref{tab:case_study}, when the provided context conflicts with common parametric knowledge about the Great Wall's length (21,196 vs.\ 8,850 kilometers), greedy decoding and COIECD fall back to the widely known but context-inconsistent estimate of 8,850 kilometers. CAD partially resolves the conflict by mentioning the updated measurement, but its response remains less complete. In contrast, CFB produces the most faithful response by correctly stating the 2012 survey result and explicitly contrasting it with the earlier estimate, thereby more closely matching the gold answer.

\paragraph{Case 2: Complementary Context}
When the input context provides additional evidence that is consistent with the model's prior knowledge, CFB can incorporate this information while maintaining coherence and fluency. In such cases, the method not only preserves the main answer but also better reflects supporting contextual details, leading to responses that are more complete, better grounded in the source, and more informative overall.

\paragraph{Case 3: Low Knowledge Conflict}
When the conflict between context and parametric knowledge is weak, CFB behaves more conservatively and tends to preserve fluent generation while still favoring source-supported content. This behavior is consistent with the goal of improving contextual faithfulness without introducing unnecessary distortion in low-conflict settings.

\section{Conclusion}

We present Context-Fidelity Boosting (CFB), a decoding-time framework for improving contextual faithfulness in language model outputs. By applying additive logit shaping to source-supported tokens, CFB improves alignment with the input context without requiring retraining. Across summarization and question answering tasks, our results show that CFB provides a simple and effective way to improve context-grounded generation while maintaining practical decoding efficiency.

\section*{Limitations}

CFB has several limitations. First, it requires access to model internals such as logits and, for the token-aware variant, signals like attention or token embeddings, which makes it difficult to apply in black-box API settings. Second, although CFB is lightweight compared with retraining-based methods, it still introduces additional decoding overhead from divergence computation and token-level relevance scoring. Third, the token-aware variant depends on the quality of local relevance estimation and is not consistently superior in high-conflict settings. Future work could explore black-box approximations, more robust token-level relevance modeling, and further reductions in decoding cost.

\bibliography{custom}

\appendix



\section{Additional Efficiency Results}
\label{app:efficiency}

We report empirical runtime per sample (in seconds) for each decoding method across three benchmark datasets. Measurements were taken on a single GPU with batch size 1.

\begin{table}[h]
\small
\centering
\begin{tabular}{lccc}
\toprule
\textbf{Method} & \textbf{CNN-DM} & \textbf{XSum} & \textbf{NQ-Synth} \\
\midrule
COIECD & 3.22 & 2.56 & 0.45 \\
Static CFB & \textbf{1.38} & \textbf{1.28} & \textbf{0.15} \\
Context-aware CFB & 2.00 & 1.86 & 0.20 \\
Token-aware CFB & 10.39 & 9.35 & 0.51 \\
\bottomrule
\end{tabular}
\caption{Average runtime per decoding sample (in seconds) across datasets.}
\label{tab:efficiency}
\vspace{-0.3cm}
\end{table}

As shown in Table~\ref{tab:efficiency}, Static CFB is the most efficient variant and is consistently faster than COIECD across all three datasets. Context-aware CFB introduces only a modest additional cost over Static CFB, reflecting the overhead of divergence computation. Token-aware CFB is substantially slower due to step-wise attention-based scoring and token-level relevance computation, but remains practical for small-batch evaluation. These results are consistent with our design goal of providing a spectrum of methods with different trade-offs between efficiency and fine-grained control.

\section{Supplementary Results on Reasoning Tasks}
\label{app:rqaqaqa}

To explore the boundary of CFB's applicability in reasoning-intensive scenarios, we conduct preliminary evaluations on multi-hop and reading comprehension QA datasets, namely HotpotQA and TriviaQA. These tasks require multi-step inference and compositional reasoning, which are not explicitly modeled by CFB.

\begin{table}[h]
\small
\centering
\begin{tabular}{lcc}
\toprule
\textbf{Method} & \textbf{HotpotQA Acc.} & \textbf{TriviaQA Acc.} \\
\midrule
Baseline (CAD) & \textbf{39.5} & 40.2 \\
Static CFB & 34.7 & 69.1 \\
Context-aware CFB & 36.8 & 70.4 \\
Token-aware CFB & 37.1 & \textbf{71.5} \\
\bottomrule
\end{tabular}
\caption{Accuracy on reasoning tasks using Llama3-8B. While not directly designed for multi-hop reasoning, CFB improves context grounding and yields strong results on TriviaQA.}
\label{tab:reasoning}
\end{table}

Table~\ref{tab:reasoning} shows that CFB is not uniformly beneficial on all reasoning-heavy tasks. On HotpotQA, CAD remains the strongest method, suggesting that simple context boosting alone is insufficient for tasks requiring stronger multi-step reasoning. In contrast, all CFB variants substantially improve over CAD on TriviaQA, with Token-aware CFB achieving the best accuracy. This pattern suggests that CFB is more effective when improved context grounding directly supports answer generation, but is less reliable when success depends primarily on complex reasoning rather than context alignment alone.

\end{document}